# AUTOMATIC SEGMENTATION OF RETINAL VASCULATURE

*Renoh Johnson Chalakkal\* and Waleed Abdulla*
University of Auckland, Auckland, New Zealand
\*rcha789@aucklanduni.ac.nz , w.abdulla@auckland.ac.nz

## ABSTRACT

Segmentation of retinal vessels from retinal fundus images is the key step in the automatic retinal image analysis. In this paper we propose a new unsupervised automatic method to segment the retinal vessels from retinal fundus images. Contrast enhancement and illumination correction are carried out through a series of image processing steps followed by adaptive histogram equalization and anisotropic diffusion filtering. This image is then converted to a gray scale using weighted scaling. The vessel edges are enhanced by boosting the detail curvelet coefficients. Optic disk pixels are removed before applying fuzzy C-mean classification to avoid the misclassification. Morphological operations and connected component analysis are applied to obtain the segmented retinal vessels. The performance of the proposed method is evaluated using DRIVE database to be able to compare with other state-of-art supervised and unsupervised methods. The overall segmentation accuracy of the proposed method is 95.18% which outperforms the other algorithms.

## 1. INTRODUCTION

Various cardiovascular and ophthalmic diseases can be diagnosed from the information that is contained in the retinal vasculature obtained from a retinal fundus image. Pathologies like diabetes, hypertension, neovascularization, macular edema and arterial sclerosis are few among them [1]. The information contained in the retinal vasculature hence can be used to predict, diagnose and analyze the progression of these pathologies in a non-invasive way [2, 3]. Our work is concentrated on using the information obtained from retinal vasculature for the diagnosis of diabetic retinopathy (DR) [4] and prevent the gradual transition of diabetic retinopathy from the non-proliferative diabetic retinopathy (NPDR) stage to proliferative diabetic retinopathy (PDR) stage. DR is a progressive disease developed as a result of blocks in the retinal vessels that supply blood and oxygen to various areas of the retina. This in turn will result in the growth of new vessels in the retina to compensate for the blocked blood vessels. These new vessels will be very delicate and fragile resulting in bleeding which can eventually lead to complete blindness [4].

The manual segmentation of retinal vasculature is an exhaustive and time consuming task. A single retinal image will take around one to two hours for the manual segmentation [5]. This is where the importance of automatic segmentation of vessels lies [6-21]. It only takes less than a minute to completely extract the retinal vasculature from the fundus images of the retina. Our work presents a novel method for the segmentation of the vessels out of the retinal fundus images.

This paper is organized as follows. In the next section we perform a thorough study about the various state-of-art methods and algorithms available for the automatic segmentation of the retinal vasculature. In section 3 we discuss the proposed algorithm in detail. Section 4 is a performance comparison between the proposed method and its counterparts. Section 5 concludes our paper with future directions of this work.

## 2. AUTOMATIC VESSEL SEGMENTATION

Automatic retinal image analysis (ARIA) is gaining importance worldwide as more and more pathologies [1-4, 22] are proved to be having some connection with the retinal image features obtained after the ARIA. Among various features that are obtained after the ARIA, the retinal vessels carries vital information about various pathologies. There are methods available in the literature which perform automatic segmentation of the retinal vessels out of the retinal fundus images. These methods could be classified into two groups in general: supervised methods and unsupervised methods. The supervised methods use prior information about the vessels in the retinal image to classify a pixel in the retinal image as a vessel or a non-vessel whereas the unsupervised methods do not use any prior information. The supervised methods rely on the accuracy of the reference standard images of extracted retinal vessels used as training set to train the algorithm.

The major drawback of supervised methods is that their performance is poor when the retinal image to be segmented is affected by untrained pathology. Many supervised methods are available in literature to segment the retinal vessels. In [14], Niemeijer has proposed a supervised method for segmenting the retinal vessels from fundus retinal images. The algorithm uses a probability map created based on the classification done using k-nearest neighbors (k-NN) [23] algorithm. The performance of this method is evaluated by testing the algorithm on DRIVE database and the accuracy is reported as 0.9416. In [15], Soares et al. discuss a supervised method based on 2-D Gabor wavelet. Pixel intensity and 2-D Gabor wavelet transform coefficients are used as the features to represent each pixel. A GMM modelling is used for the classification purpose. The accuracy reported on database DRIVE is 0.9466.This method suffers a setback when the images are of non-uniform illumination which results in false classification of the pixels belonging to the optic disk border, exudates and other features as the vessel pixel. You et al. [13] proposed a method based on radial projections and SVM classifier. This method works well in the case of healthy retinal images whereas for the diseased images it has a poor accuracy. The average accuracy of this method is 0.9434 for DRIVE database. Staal [16] has discussed a method based on the extraction of ridges from the retinal images. These ridges are considered to be the pointers to the vessels. k-NN classifier is used for the classification after extracting the features. The reported accuracy is 0.9516. [11, 17-20] discuss various other supervised methods available for the retinal vessel segmentation. The performance of these methods are tabulated in section 4.

Zana et al. [21] discuss an unsupervised method based on morphology processing. This method depends on the specific characteristic of the structuring elements selected for the morphological processing. The performance of this method is evaluated in terms of true positive rate and false positive rate. The average accuracy for this method is reported to be 0.9377 on the DRIVE database. In [8], another unsupervised method is discussed in which differential filters are used to extract the vessel centerlines. Then morphological operations are used to fill in the regions. This method reports an average accuracy of 0.9452 on DRIVE database. The drawback here is that, when the vessel centerlines are missed, the retinal vessels are also missed. Hoover et al. [24], discuss a method based on matched filtering to extract the vessels from the retinal images. This method uses the matched filter method proposed by Chaudhuri et al. [25] to obtain the matched filter response (MFR) image. Then this MFR image is analyzed in pieces using this method using both local and region based properties. They have reported an average accuracy of 0.9267. In [6], Fraz et al. have discussed a method based on the vessel centerlines similar to [8]. The only difference is that they combine the vessel centerline detection with the bit plane slicing using morphological operations. They have reported an average accuracy of 0.9430 on DRIVE database. In [10], Lam et al. presented a method based on the vessel concavity. This method was able to extract the vessels from both healthy and pathological retinal images. Other unsupervised methods are discussed in [7, 9, 12] whose performance are reported in section 4.

## 3. THE PROPOSED METHOD

We have developed an unsupervised method for segmenting the retinal vessels out of the digital retinal fundus images. The entire algorithm consists of four major stages: contrast enhancement, edge enhancement, optic disk (OD) removal and vessel segmentation and the post-processing. These steps are discussed in detail in the following subsection.

### 3.1 Contrast enhancement

The input image is resized to $500 \times 500$ in order to match the coding requirements. The resized image ($I_{orig}$) is converted to the CIELab color space [26-28] consisting of three components, the luminosity layer '$L$', and the chromaticity layers '$a$' and '$b$'. The '$L$' layer is used in the further steps of this algorithm as the '$L$' layer of the retinal fundus image provides better contrast and more structural information compared to the chromaticity layers '$a$' and '$b$'. In order to further increase the contrast, adaptive histogram equalization is performed on the '$L$' layer [29]. Since the adaptive histogram equalization also increases the noise in the retinal image, anisotropic diffusion filtering [30, 31] is carried out to suppress the enhanced noise. After the filtering the resulting image is converted back to the RGB color space ($I_{RGB}$). The difference in contrast between the original RGB image ($I_{orig}$) and $I_{RGB}$ is shown in Fig.1.(a) and Fig.1.(b) respectively. We performed a weighted scaling [32] on the three layers of $I_{RGB}$ according to Eq.3.1 to convert it to a gray scale image ($I_{weighted}$).

$$I_{weighted} = 0.299 * I_{red} + 0.587 * I_{green} + 0.114 * I_{blue} \quad (3.1)$$

where $I_{red}$, $I_{green}$ and $I_{blue}$ are the three layers of the $I_{RGB}$. The contrast is much better in the resulting image after this step compared to the green channel of $I_{orig}$ used in many state of art algorithms [14, 17, 33, 34]. One of the main factors that makes our proposed method superior to other state-of-art methods is this contrast improvement of the retinal image achieved using our proposed method. This is evident from Fig.1.(c) and Fig.1.(d), which show the difference in the contrast between the green channel of $I_{orig}$ and the image obtained by our proposed algorithm ($I_{weighted}$). The retinal vessels are more highlighted in the weighted scale image compared to the green channel of the original image.

### 3.2 Edge enhancement

A retinal fundus mask is applied on $I_{weighted}$ to extract the region of interest. In order to enhance the retinal vessel edges, we applied curvelet transform [35] on the image.

Curvelet transform is a multiscale transform and the two important properties (anisotropy-scaling law and the directionality) of this transform, makes it the best possible transform to be applied on the retinal image as it can efficiently handle the singularities present in it and can enhance the vessel edges. Many state of art methods [33, 34, 36-39], uses curvelet transform to process the retinal image prior to the extraction of the vessels. After obtaining the curvelet coefficients of the retinal image, we unlike the other methods [33, 34, 36-39], suppress the coefficients corresponding to the coarse approximation of the images to zero and multiply the remaining subband coefficients by a suitable amplification factor '$\kappa$'. This process makes the edges of the retinal vessels more evident. The coarse coefficients are basically interpreted as the background. Then inverse curvelet transform is applied to obtain the transformed image which is then multiplied with the weighted scale image, $I_{weighted}$ to obtain the vessel edge enhanced image, $I_{edge}$ as shown in Fig.1.(e).

### 3.3 OD removal and vessel segmentation

OD pixels can be wrongly classified as vessels. This can seriously affect the performance of any vessel detection algorithm. Hence we removed the OD from the input retinal image using a series of image processing steps. The weighted scale image, $I_{weighted}$ is low pass filtered using a median filter of size $20 \times 20$ to obtain a background image, $I_{bg}$ (Eq. 3.2). This background image is then subtracted from the weighted scale image (Eq. 3.3) to set a thresholding intensity, $I_t$ having intensity values less than zero (Eq. 3.4). Finally the image with OD removed, $I_{out}$ is introduced using (Eq. 3.5), where '$c$' represents the complement taken.

The resulting image is shown in Fig.1. (f). From the figure it is clear that that the OD pixels are well differentiated from the vessel pixels which is a must for the segmentation step.

$$I_{bg} = LPF(I_{weighted}) \qquad (3.2)$$

$$I_{new} = I_{weighted} - I_{bg} \qquad (3.3)$$

$$I_t = (I_{new} \leq 0) \qquad (3.4)$$

$$I_{out} = (I_{edge} - I_t)^c \qquad (3.5)$$

The vessels are segmented out from the retinal image, $I_{out}$ by applying fuzzy C-mean (FCM) clustering algorithm [40, 41]. The FCM assigns a particular degree of membership for each class in the image. The pixels that have close membership values form new clusters. The input retinal image is classified into three classes using FCM: vessel pixels, exudate pixels (including OD pixels) and the background pixels. The objective function $Q$ of the FCM is given in Eq. 3.6.

$$Q = \sum_{i=1}^{n}\sum_{j=1}^{C} u_{ij}^m \|v_i - \mu_j\|^2 \qquad 1 \leq m \leq \infty \qquad (3.6)$$

where $n$ is the number of samples and $C$ is the number of clusters, $\mu_j$ is the d-dimension center of the cluster $j$, $m \in [1, \infty]$ is a weighting exponent, $v_i$ is the $i^{th}$ d-dimensional measured data, and $\| * \|$ is a norm expressing the distance between the measured data and the cluster center. In the proposed method we have used $C = 3$ and $m = 2$. The matrix $u$ is the membership matrix, which satisfies the condition:

$$\sum_{i=1}^{C} u_{ij} = 1, \forall j = 1, \ldots, n \qquad (3.7)$$

With each iteration the memberships and the cluster centers are updated using Eq. 3.8

$$u_{ij}^m = \frac{1}{\sum_{k=1}^{n}\left(\frac{\|v_i - \mu_i\|}{\|v_i - \mu_k\|}\right)^{\frac{2}{m-1}}} \quad and \quad \mu_j = \frac{\sum_{i=1}^{n} u_{ij} v_i}{\sum_{i=1}^{n} u_{ij}} \qquad (3.8)$$

Vessels extracted out after this step is shown in Fig.1.(g)

### 3.4 Post processing

For improving the accuracy of the proposed method, the output binary image obtained after the FCM classification is processed through a series of post processing steps involving morphological processing and connected component analysis (CCA) [42]. Morphological operations like dilation and bridging [43] are being carried out to connect the broken vessels. CCA is carried out on the morphologically processed image to further improve the accuracy of the proposed method. CCA eliminates the wrongly detected pixels that belong to exudates and the macular regions of the retinal image. The final output of our proposed algorithm is reflected in Fig.1.(h).

### 3.5 Programming environment and database

We have used MATLAB R2016a version for the programming purpose. We used a computer with system's configuration of Intel core i7, 3.60 GHz with 8 GB RAM.

DRIVE database [44] is used for evaluating the performance of the proposed method. DRIVE database consists of 40 retinal fundus images. These images are captured using a Canon CR5 non-mydriatic 3-CCD camera. All the images in this database have 45° field of view. These 40 images are divided into testing and training sets with 20 images each. Most of the state-of-art retinal vessel segmentation algorithms [7-9, 11-15, 17-21] reported their performance based on the DRIVE database.

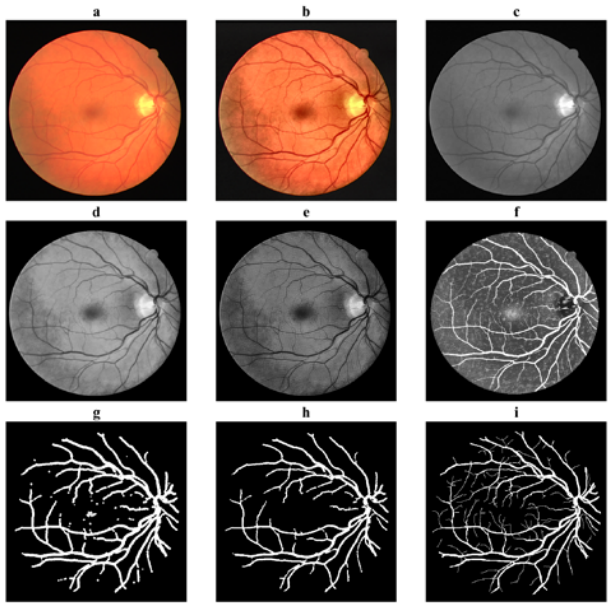

**Fig.1.** (a) original retinal fundus image ('02_test.tif' from DRIVE database), (b) contrast enhanced image, (c) green channel image of original image, (d) weighted scale image, (e) vessel edge enhanced image, (f) OD removed vessel enhanced image, (g) segmented vessels after applying FCM, (h) final result of vessel segmentation (after morphological processing and CCA), (i) manually segmented reference image ('02_manual1.gif') from DRIVE database.

## 4. RESULTS AND DISCUSSION

The performance of the proposed method is evaluated in terms of sensitivity, specificity and accuracy. Since the retinal vessel classification results in two class of pixels, we have four possibilities for the classification done viz. true positive (TP) and true negative (TN) which are the values corresponding to correct classifications made whereas corresponding to the misclassifications made, we have false positive (FP) and false negative (FN). These four measures are used to calculate the sensitivity (SN), specificity (SP) and accuracy (Acc). Since most of the state-of-art algorithms have used only the first 20 test images from the DRIVE database, we have mentioned the result obtained for the first 20 images from the DRIVE database to be able to compare.

From Table.1, it can be noticed that we achieved better performance compared to the other state-of-art methods. The accuracy of the proposed method is 0.9518, which is the highest among the unsupervised methods [6-10, 12, 21] and comparable to the supervised methods [17-18]. The improved accuracy achieved in [17-18] is due to compromising the sensitivity. The overall performance of the proposed algorithm is superior to most of the state-of-art methods including the supervised ones. The overall accuracy of the proposed method, sensitivity and specificity (including all 40 images of DRIVE) is 0.9510, 0.7382 and 0.9761 respectively.

**Table 1**
Comparison of vessel extraction results on DRIVE* database

| No | Methods | SN | SP | Acc (%) |
|---|---|---|---|---|
| 1 | Zana et al. [21] | 0.6971 | – | 0.9377 |
| 2 | Jiang et al. [7] | – | – | 0.9212 |
| 3 | Mendonca et al. [8] | 0.7344 | 0.9764 | 0.9452 |
| 4 | Al-Diri et al. [9] | 0.7282 | 0.9551 | – |
| 5 | Lam et al. [10] | – | – | 0.9472 |
| 7 | Fraz et al. [6] | 0.7152 | 0.9759 | 0.9430 |
| 8 | Zhao et al. [12] | 0.7354 | 0.9789 | 0.9477 |
| 9 | **Proposed method** | *0.7386* | *0.9769* | *0.9518* |
| 10 | You et al. [13] | 0.7410 | 0.9751 | 0.9434 |
| 11 | Niemeijer et al. [14] | – | – | 0.9416 |
| 12 | Soares et al. [15] | 0.7332 | 0.9782 | 0.9461 |
| 13 | Staal et al. [16] | – | – | 0.9441 |
| 14 | Ricci et al. [17] | – | – | 0.9595 |
| 15 | Lupascu et al. [18] | 0.7200 | – | 0.9597 |
| 16 | Marin et al. [19] | 0.7067 | 0.9801 | 0.9452 |
| 17 | Fraz et al. [11] | 0.7406 | 0.9807 | 0.9480 |
| 18 | Vega et al. [20] | 0.7444 | 0.9600 | 0.9412 |

*test set (first 20 images) were used for the performance comparison

## 5. CONCLUSION AND FURTHER WORKS

This paper presents an unsupervised method for the segmentation of retinal blood vessels. The proposed method achieves the highest sensitivity (0.7386) and accuracy (0.9518) among the unsupervised methods [6-10, 12, 21]. One of the main contributions of this paper is the method proposed for the contrast enhancement of the digital retinal fundus images. This helps the proposed method to perform well on retinal images with uneven contrast and illumination which is advantageous over the common state-of-art methods. The enhancement of vessel edges using curvelet transform and OD removal helps in reducing the misclassification during the FCM classification. This helps the proposed method to achieve better segmentation accuracy while segmenting vessels from the retinal images affected by various pathologies. The accuracy of proposed method is comparable to supervised methods [17-18] which reports the highest accuracy. Even the supervised method that reports the highest accuracy of 0.9597 [18] has a sensitivity of 0.7200 which is less than the sensitivity (0.7386) of our proposed method. Our next immediate task is to improve the proposed algorithm to detect the delicate and thin vessels in the retinal images which can further increase the overall segmentation accuracy and could help in the automatic DR diagnosis.